\documentclass[10pt,twocolumn,letterpaper]{article}
\pdfoutput=1

\usepackage{enumitem}
\usepackage{cvpr}
\usepackage{times}
\usepackage{epsfig}
\usepackage{graphicx}
\usepackage{amsmath}
\usepackage{amssymb}
\usepackage{xspace}
\usepackage{multirow}
\usepackage{blindtext}
\usepackage{arydshln}
\usepackage{algorithm} 
\usepackage{algpseudocode} 

\newcommand{\mypar}[1]{\vspace{-2mm}\paragraph{#1}} 

\newcommand{\lidar}{lidar\xspace} 
\newcommand{\bev}{bird's-eye view\xspace} 

\usepackage[usenames, dvipsnames]{color} 
\definecolor{purple}{rgb}{0.5, 0.0, 0.5}
\definecolor{orange}{rgb}{1, 0.65, 0}
\definecolor{lightgreen}{rgb}{0.68, 1, 0.18}
\definecolor{darkgreen}{rgb}{0, 0.7, 0}
\definecolor{darkred}{rgb}{0.6, 0, 0}
\definecolor{brown}{rgb}{0.64, 0.16, 0.16}
\iftrue 
  \newcommand{\bassam}[1]{\noindent}
  \newcommand{\oscar}[1]{\noindent}
  \newcommand{\alex}[1]{\noindent}
  \newcommand{\sourabh}[1]{\noindent}
  \newcommand{\done}[1]{\noindent}
  \newcommand{\todo}[1]{\noindent}
\else
  \newcommand{\bassam}[1]{\textcolor{blue}{\bf [BH: #1]}}
  \newcommand{\oscar}[1]{\textcolor{orange}{\bf [OB: #1]}}
  \newcommand{\alex}[1]{\textcolor{purple}{\bf [AL: #1]}}
  \newcommand{\sourabh}[1]{\textcolor{brown}{\bf [SV: #1]}}
  \newcommand{\done}[1]{\textcolor{darkgreen}{\bf [Done: #1]}}
  \newcommand{\todo}[1]{\textcolor{red}{\bf [Todo: #1]}}
\fi
\usepackage[normalem]{ulem} 

\newcommand{\figref}[1]{Figure \ref{#1}}
\newcommand{\tableref}[1]{Table \ref{#1}}

\newcommand{\squeeze}{\vspace{-0mm}}
\newcommand{\greenbf}[1]{\textcolor{darkgreen}{\bf #1}}
\newcommand{\green}[1]{\textcolor{darkgreen}{#1}}
\newcommand{\red}[1]{\textcolor{red}{#1}}

\usepackage[pagebackref=true,breaklinks=true,letterpaper=true,colorlinks,bookmarks=false]{hyperref}

\cvprfinalcopy 

\def\cvprPaperID{7258} 

\ifcvprfinal\fi
\begin{document}

\title{PointPainting: Sequential Fusion for 3D Object Detection}

\author{
Sourabh Vora  \hspace{10mm} Alex H. Lang  \hspace{10mm} Bassam Helou  \hspace{10mm} Oscar Beijbom \\
nuTonomy: an Aptiv Company\\
{\tt\small \{sourabh, alex, bassam, oscar\}@nutonomy.com}
}

\maketitle
\thispagestyle{empty}

\begin{abstract}
Camera and \lidar are important sensor modalities for robotics in general and self-driving cars in particular.
The sensors provide complementary information offering an opportunity for tight sensor-fusion.
Surprisingly, lidar-only methods outperform fusion methods on the main benchmark datasets, suggesting a gap in the literature.
In this work, we propose PointPainting: a sequential fusion method to fill this gap.
PointPainting works by projecting lidar points into the output of an image-only semantic segmentation network and appending the class scores to each point.
The appended (painted) point cloud can then be fed to any lidar-only method.
Experiments show large improvements on three different state-of-the art methods, Point-RCNN, VoxelNet and PointPillars on the KITTI and nuScenes datasets.
The painted version of PointRCNN represents a new state of the art on the KITTI leaderboard for the \bev detection task.
In ablation, we study how the effects of Painting depends on the quality and format of the semantic segmentation output, and demonstrate how latency can be minimized through pipelining.
\end{abstract}

\begin{figure}
\begin{center}
\includegraphics[width = \columnwidth]{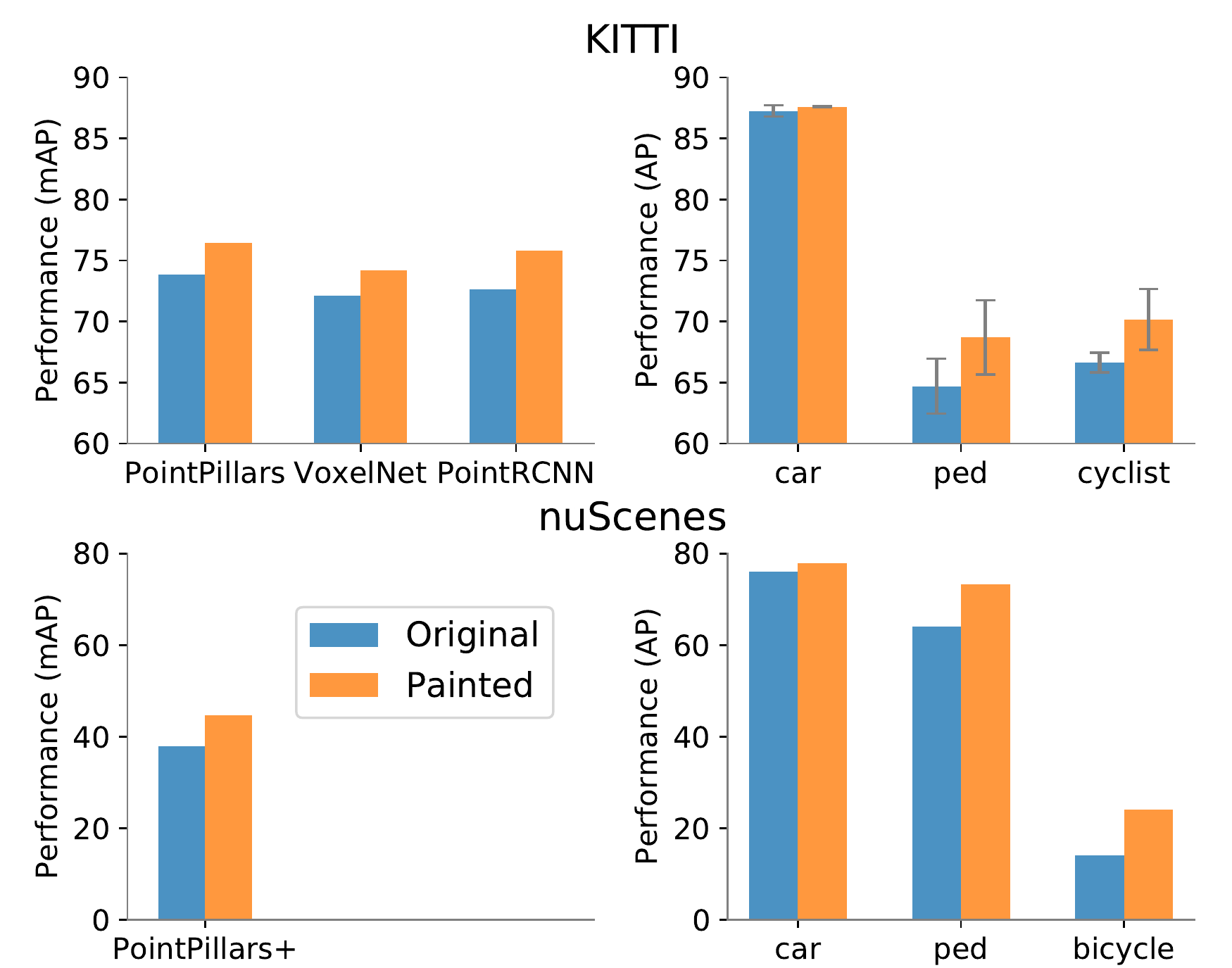}
\end{center}
\vspace{-2mm}
\caption{
PointPainting is a general fusion method that be can used with any lidar detection network. 
Top left: PointPillars\cite{pointpillars}, VoxelNet~\cite{voxelnet, second}, and PointRCNN~\cite{pointrcnn} on the KITTI~\cite{kitti} \bev \emph{val} set (\tableref{table:point_delta}). 
The painted version of PointRCNN is state of the art on the KITTI \emph{test} set outperforming all published fusion and lidar-only methods (\tableref{table:res_bev}). 
Top right: improvements are larger for the harder pedestrian (ped.) and cyclist classes. 
Error bars indicate std. across methods.
Bottom left: PointPillars+ evaluated on the nuScenes~\cite{nuscenes} \emph{test} set. 
The painted version of PointPillars+ improves all 10 classes for a total boost of 6.3 mAP (\tableref{table:nuscenes_per_class}).
Bottom right: selected class improvements for Painted PointPillars+ show the challenging bicycle class has the largest gains.
}
\label{fig:point_painting_delta}
\end{figure}

\squeeze
\section{Introduction} \label{sec:intro}
\squeeze

\begin{figure*}
\begin{center}
\includegraphics[width = 1.0\textwidth]{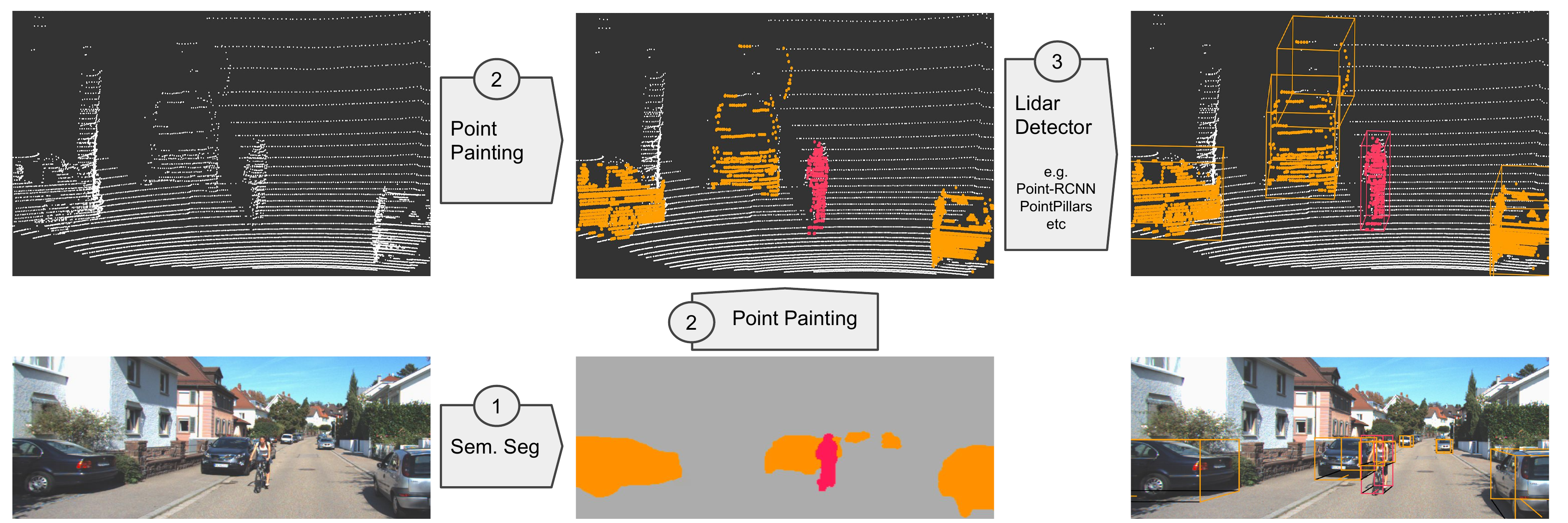}
\end{center}
\vspace{-3mm}
\caption{PointPainting overview.
The PointPainting architecture consists of three main stages: (1) image based semantics network, (2) fusion (painting), and (3) lidar based detector.
In the first step, the images are passed through a semantic segmentation network obtaining pixelwise segmentation scores.
In the second stage, the \lidar points are projected into the segmentation mask and decorated with the scores obtained in the earlier step.
Finally, a lidar based object detector can be used on this decorated (painted) point cloud to obtain 3D detections.
}
\label{fig:network}
\label{fig:pointpainting}
\end{figure*}

Driven partially by the interest in self-driving vehicles, significant research effort has been devoted to 3D object detection.
In this work we consider the problem of fusing a lidar point cloud with an RGB image.
The point cloud provides a very accurate range view, but with low resolution and texture information.
The image, on the other hand, has an inherent depth ambiguity but offers fine-grained texture and color information.
This offers the compelling research opportunity of how to design a detector which utilizes the best of two worlds.

Early work on KITTI~\cite{kitti} such as MV3D~\cite{mv3d} and AVOD~\cite{avod} proposed multi-view fusion pipelines to exploit these synergies.
However recent detectors such as PointPillars~\cite{pointpillars}, VoxelNet~\cite{voxelnet, second} and STD~\cite{yang2019std} use only \lidar and still significantly outperform these methods.
Indeed, despite recent fusion research~\cite{frustum, sparsepool, liang2019multi, you2019pseudo}, the top methods on the popular KITTI leaderboard~\cite{kitti} are \lidar only.
Does this mean lidar makes vision redundant for 3D object detection?

\begin{figure}
\begin{center}
\includegraphics[width=0.9\columnwidth]{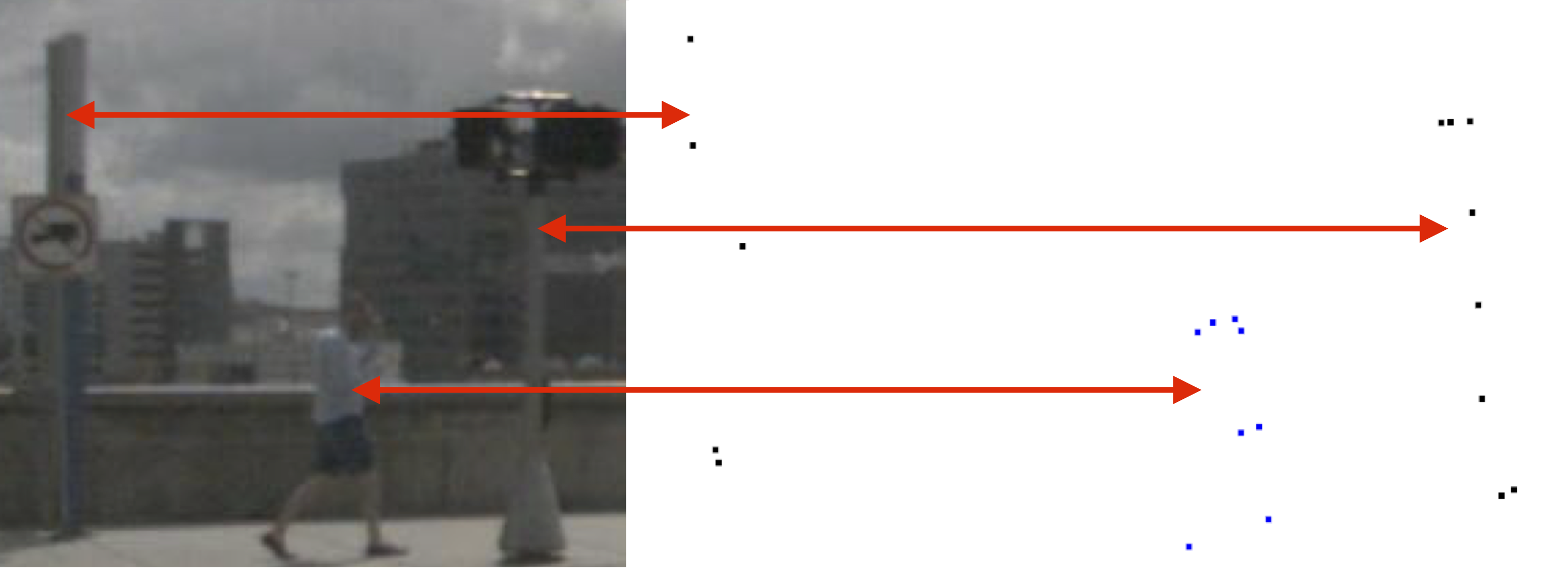}
\end{center}
\vspace{-3.3mm}
\caption{
Example scene from the nuScenes~\cite{nuscenes} dataset.
The pedestrian and pole are 25 meters away from the ego vehicle.
At this distance the two objects appears very similar in the point cloud.
The proposed PointPainting method would add semantics from the image making the lidar detection task easier.
}
\label{fig:ped_vs_pole}
\end{figure}

The answer, surely, must be no.
Consider the example in Fig. \ref{fig:ped_vs_pole}, where the pedestrian and signpost are clearly visible in the image, yet look more or less identical in the lidar modality.
Surely vision based semantic information should be useful to improve detection of such objects.
Also, by first principle, adding more information should at the minimum yield the \emph{same} result, not \emph{worse}.
So why has it been so difficult?
One reason is due to viewpoint misalignment.

While both sensors are natively captured in the range-view, most state of the art methods such as PointPillars~\cite{pointpillars} or STD~\cite{yang2019std} use convolutions in the \bev.
This view has several advantages including lack of scale ambiguity and minimal occlusions.
It also does not suffer from the depth-blurring effect which occurs with applying 2D convolutions to the range view~\cite{pseudo_lidar}.
As a result, \bev methods outperform top range-view methods, such as LaserNet~\cite{lasernet, lasernet++}, on the KITTI leaderboard.
However, while a \lidar point cloud can trivially be converted to \bev, it is much more difficult to do so with an image.

Hence, a core challenge of sensor fusion network design lies in consolidating the lidar \bev with the camera view.
Previous methods can be grouped into four categories: object-centric fusion, continuous feature fusion, explicit transform and detection seeding.

Object-centric fusion, pioneered by MV3D~\cite{mv3d} and AVOD~\cite{avod}, is the most obvious choice for a two-stage architecture.
Here, the modalities have different backbones, one in each view, and fusion happens at the object proposal level by applying roi-pooling in each modality from a shared set of 3D proposals.
This allows for end-to-end optimization but tends to be slow and cumbersome.

A second family of methods applies ``continuous feature fusion'' to allow feature information to be shared across all strides of the image and lidar backbones~\cite{contfuse, sparsepool}.
These methods can be used with single-state detection designs but require a mapping to be calculated, \emph{a priori}, for each sample, from the point-cloud to the image.
One subtle but important draw-back of this family of methods is ``feature-blurring''.
This occurs since each feature vector from the \bev corresponds to multiple pixels in the image-view, and vice versa.
ContFuse~\cite{contfuse} proposes a sophisticated method based on kNN, bilinear interpolation and a learned MLP to remedy this, but the core problem persists.

A third family of methods attempts to explicitly transform the image to a \bev representation~\cite{oft} and do the fusion there~\cite{pseudo_lidar, weng2019monocular, you2019pseudo}.
Some of the most promising \emph{image-only} methods use this idea of first creating an artificial point cloud from the image and then proceeding in the \bev~\cite{pseudo_lidar, weng2019monocular}.
Subsequent work attempts fusion based on this idea, but the performance falls short of state of the art~\cite{you2019pseudo}, and requires several expensive steps of processing to build the pseudo-point cloud.

\begin{table*}[]
\small
\center
\begin{tabular}{| c | c || c | c | c || c | c | c || c | c | c |}
\hline
\multirow{2}{*}{Method}  & mAP  & \multicolumn{3}{|c||}{Car} & \multicolumn{3}{|c||}{Pedestrian} & \multicolumn{3}{|c|}{Cyclist} \\ \cline{2-11}
& Mod. & Easy   & Mod.   & Hard   & Easy      & Mod.     & Hard     & Easy     & Mod.    & Hard    \\ \hline \hline
PointPillars~\cite{pointpillars}    		& 73.78    		& 90.09  & 87.57  & 86.03     		& 71.97  & 67.84  & 62.41             			& 85.74  & 65.92  & 62.40      \\
Painted PointPillars 				& 76.27    		& 90.01  & 87.65  & 85.56        		& 77.25  & 72.41  & 67.53		        			& 81.72  & 68.76  & 63.99			\\ \hdashline
Delta 		 				& \green{+2.50}    		& \red{-0.08}     & \green{0.08}	   & \red{-0.47}          	& \greenbf{+5.28}  & \greenbf{+4.57}  & \greenbf{+5.12}		& \red{-4.02}  & \green{+2.84}  & \green{+1.59}			\\ \hline \hline
VoxelNet~\cite{voxelnet,second}       & 71.83   	 	& 89.87  & 87.29  & 86.30    		& 70.08  & 62.44  & 55.02           			& 85.48  & 65.77  & 58.97        		\\
Painted VoxelNet     				& 73.55    		& 90.05  & 87.51  & 86.66        		& 73.16  & 65.05  & 57.33					& 87.46  & 68.08  & 65.59			\\ \hdashline
Delta			 				& \green{+1.71}    		& \green{+0.18}  & \green{+0.22}  & \green{+0.36}     & \green{+3.08}  & \green{+2.61}  & \green{+2.31}					& \greenbf{+1.98}  & \green{+2.31}  & \green{+6.62}			\\ \hline \hline
PointRCNN~\cite{pointrcnn}      	& 72.42    		& 89.78  & 86.19  & 85.02     		& 68.37  & 63.49  & 57.89             			& 84.65  & 67.59  & 63.06        		\\
Painted PointRCNN  			& 75.80   		& 90.19  & 87.64  & 86.71          	& 72.65  & 66.06  & 61.24					& 86.33  & 73.69  & 70.17  			\\ \hdashline
Delta 		 				& \greenbf{+3.37} &  \greenbf{+0.41}  & \greenbf{+1.45}  & \greenbf{+1.69}          		& \green{+4.28}  & \green{+2.57}  & \green{+3.35}					& \green{+1.68}  & \greenbf{+6.10} & \greenbf{+7.11}		\\ \hline
\end{tabular}
\vspace{1mm}
\caption{
PointPainting applied to state of the art \lidar based object detectors.
All lidar methods show an improvement in \bev (BEV) mean average precision (mAP) of car, pedestrian, and cyclist on KITTI \emph{val} set, moderate split.
The corresponding 3D results are included in \tableref{table:point_delta_3d} in the Supplementary Material where we observe a similar improvement.
}
\label{table:point_delta}
\end{table*}

A fourth family of methods uses detection seeding.
There, semantics are extracted from an image \emph{a priori} and used to seed detection in the point cloud. Frustrum PointNet~\cite{frustum} and ConvNet~\cite{fconvnet} use the 2D detections to limit the search space inside the frustum while IPOD~\cite{ipod} uses semantic segmentation outputs to seed the 3D proposal.
This improves precision, but imposes an upper bound on recall.

The recent work of Liang et. al~\cite{liang2019multi} tried combining several of these concepts.
Results are not disclosed for all classes, but the method is outperformed on the car class by the top lidar-only method STD~\cite{yang2019std} (\tableref{table:res_bev}).

In this work we propose PointPainting: a simple yet effective sequential fusion method.
Each \lidar point is projected into the output of an image semantic segmentation network and the channel-wise activations are concatenated to the intensity measurement of each \lidar point.
The concatenated (painted) lidar points can then be used in any \lidar detection method, whether \bev~\cite{pointpillars, second, pointrcnn, voxelnet, pixor, yang2019std} or front-view~\cite{fcl, lasernet}.
PointPainting addresses the shortcomings of the previous fusion concepts:
it does not add any restrictions on the 3D detection architecture; it does not suffer from feature or depth blurring; it does not require a pseudo-point cloud to be computed, and it does not limit the maximum recall.

Note that for lidar detection methods that operate directly on the raw point cloud~\cite{pointpillars, voxelnet, lasernet, pointrcnn}, PointPainting requires minimal network adaptations such as changing the number of channels dedicated to reading the point cloud.
For methods using hand-coded features~\cite{pixor, complexyolo}, some extra work is required to modify the feature encoder.

PointPainting is sequential by design which means that it is not always possible to optimize, end-to-end, for the final task of 3D detection.
In theory, this implies sub-optimality in terms of performance.
Empirically, however, PointPainting is more effective than all other proposed fusion methods.
Further, a sequential approach has other advantages:
(1) semantic segmentation of an image is often a useful stand-alone intermediate product, and
(2) in a real-time 3D detection system, latency can be reduced by pipelining the image and lidar networks such that the lidar points are decorated with the semantics from the previous image.
We show in ablation that such pipelining does not affect performance.

We implement PointPainting with three state of the art \lidar-only methods that have public code: PointPillars~\cite{pointpillars}, VoxelNet (SECOND)~\cite{voxelnet, second}, and PointRCNN~\cite{pointrcnn}. PointPainting consistently improved results (\figref{fig:point_painting_delta}) and indeed, the painted version of PointRCNN achieves state of the art on the KITTI leaderboard (\tableref{table:res_bev}).
We also show a significant improvement of 6.3 mAP (\tableref{table:nuscenes_test}) for Painted PointPillars+ on nuScenes~\cite{nuscenes}.

\mypar{Contributions.}
Our main contribution is a novel fusion method, PointPainting, that augments the point cloud with image semantics.
Through extensive experimentation we show that PointPainting is:
\begin{itemize}[noitemsep,topsep=0pt,parsep=0pt,partopsep=0pt]
\setlength\itemsep{.1mm}
\item \textbf{general} -- achieving significant improvements when used with 3 top lidar-only methods on the KITTI and nuScenes benchmarks;
\item \textbf{accurate} -- the painted version of PointRCNN achieves state of the art on the KITTI benchmark;
\item \textbf{robust} -- the painted versions of PointRCNN and PointPillars improved performance on \emph{all classes} on the KITTI and nuScenes \emph{test} sets, respectively.
\item \textbf{fast} -- low latency fusion can be achieved by pipelining the image and lidar processing steps.
\end{itemize}

\begin{figure*}
\begin{center}
\includegraphics[width = 1.0\textwidth]{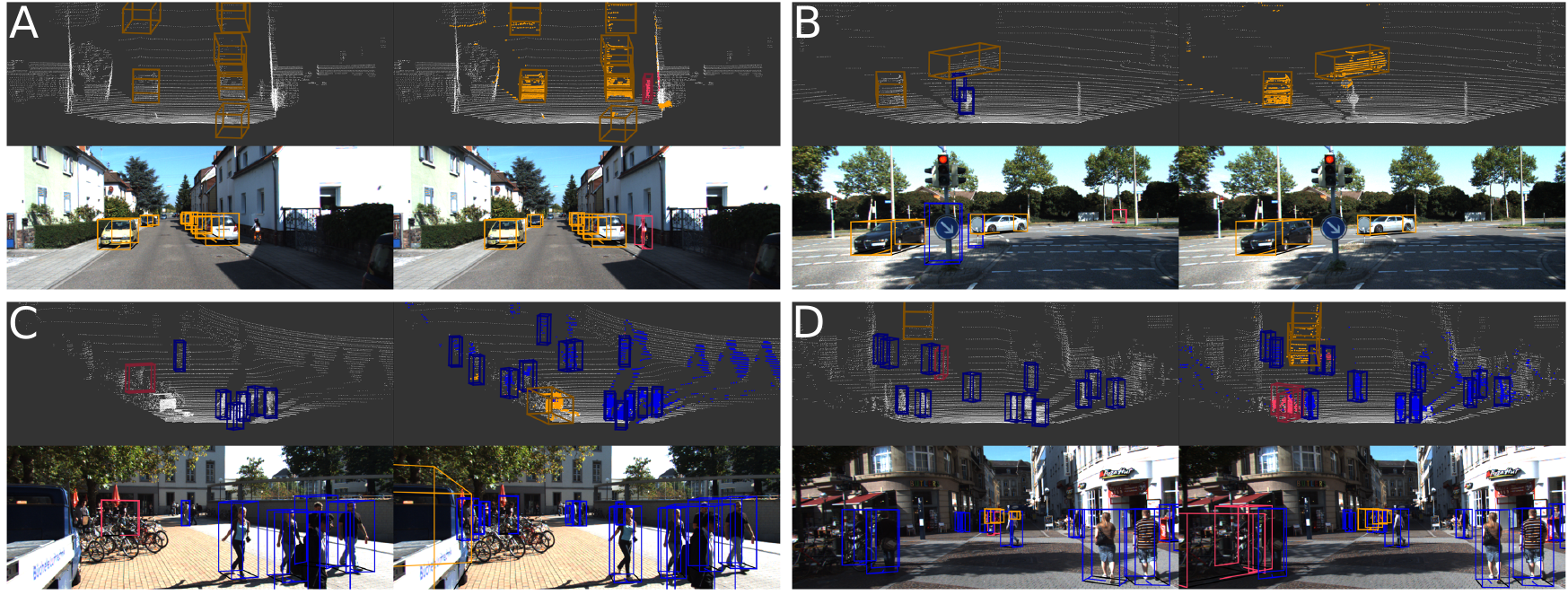}
\end{center}
\vspace{-3.3mm}
\caption{Qualitative analysis of KITTI results.
We created four different comparison figures.
For each comparison, the upper left is the original point cloud, while the upper right is the painted point cloud with the segmentation outputs used to color car (orange), cyclist (red) and pedestrian (blue) points.
PointPillars / Painted PointPillars predicted 3D bounding boxes are displayed on the both the input point cloud (upper left / right) and projected into the image (lower left / right).
The orientation of boxes is shown by a line connecting the bottom center to the front of the box.
}
\label{fig:kitti_visualize}
\end{figure*}


\section{PointPainting Architecture} \label{sec:network}
\squeeze

The PointPainting architecture accepts point clouds and images as input and estimates oriented 3D boxes.
It consists of three main stages (Fig.~\ref{fig:pointpainting}).
(1) Semantic Segmentation: an image based sem. seg. network which computes the pixel wise segmentation scores.
(2) Fusion: \lidar points are painted with sem. seg. scores.
(3) 3D Object Detection: a lidar based 3D detection network.

\subsection{Image Based Semantics Network}
\label{sec:image-segmentation-network}
The image sem. seg. network takes in an input image and outputs per pixel class scores.
These scores serve as compact summarized features of the image.
There are several key advantages of using sem. seg. in a fusion pipeline.
First, sem. seg. is an easier task than 3D object detection since segmentation only requires local, per pixel classification, while object detection requires 3D localization and classification.
Networks that perform sem. seg. are easier to train and are also amenable to perform fast inference.
Second, rapid advances are being made in sem. seg.~\cite{cheng2019panopticdeeplab,zhu2019improving}, which allows PointPainting to benefit from advances in both segmentation and 3D object detection.
Finally, in a robotics or autonomous vehicle system, sem. seg. outputs are useful independent outputs for tasks like free-space estimation.

In this paper, the segmentation scores for our KITTI experiments are generated from DeepLabv3+~\cite{deeplabv3plus,zhu2019improving,reda2018sdc}, while for nuScenes experiments we trained a custom, lighter, network.
However, we note that PointPainting is agnostic to the image segmentation network design.

\subsection{PointPainting}
\label{sec:projection}

\begin{algorithm}
    \begin{algorithmic}
        \caption{PointPainting(L, S, T, M)} \label{algorithm:painting}
        \State \textbf{Inputs:}
        \State Lidar point cloud $L \in \mathbb{R}^{N, D}$ with $N$ points and $D \geq 3$.
        \State Segmentation scores $S \in \mathbb{R}^{W, H, C}$ with $C$ classes.
        \State Homogenous transformation matrix $T \in \mathbb{R}^{4,4}$.
        \State Camera matrix $ M \in \mathbb{R}^{3,4}$.
        \State
        \State \textbf{Output:}
        \State Painted lidar points $P \in \mathbb{R}^{N, D+C}$
    	\State
        \For{$\vec{l}$ $\in L$}
              \State $\vec{l}_{\rm{image}} = \textrm{PROJECT}(M, T, \vec{l}_{xyz}$)   	 \Comment $\vec{l}_{\rm{image}} \in \mathbb{R}^2$
              \State $\vec{s} = S[\vec{l}_{\rm{image}}[0], \vec{l}_{\rm{image}}[1], :]$         	\Comment $\vec{s} \in \mathbb{R}^C$
              \State $\vec{p} = \mbox{Concatenate} (\vec{l}, \vec{s} )$					\Comment $\vec{p} \in \mathbb{R}^{D+C}$
         \EndFor
    \end{algorithmic}
\end{algorithm}

\begin{table*}[]
\small
\center
\begin{tabular}{| c | c | c || c | c | c || c | c | c || c | c | c |}
\hline
\multirow{2}{*}{Method} & \multirow{2}{*}{Modality}  & mAP  & \multicolumn{3}{|c||}{Car} & \multicolumn{3}{|c||}{Pedestrian} & \multicolumn{3}{|c|}{Cyclist} \\ \cline{3-12}
                        				&                      & Mod. & Easy   & Mod.   & Hard   & Easy      & Mod.     & Hard     & Easy     & Mod.    & Hard    \\ \hline \hline
MV3D\cite{mv3d}     			& L \& I             & N/A  	& 86.62  & 78.93  & 69.80  & N/A       & N/A      & N/A      & N/A      & N/A     & N/A     \\
AVOD-FPN\cite{avod}       	& L \& I             & 64.07   & 90.99  & 84.82  & 79.62  & 58.49     & \textbf{50.32}    & \textbf{46.98}    & 69.39    & 57.12   & 51.09   \\
IPOD\cite{ipod}           		& L \& I             &  64.60  	& 89.64  & 84.62  & 79.96  & \textbf{60.88}     & 49.79    & 45.43    & 78.19      & 59.40   & 51.38   \\
F-PointNet\cite{frustum} 		& L \& I            	& 65.20   & 91.17  & 84.67  & 74.77  & 57.13     & 49.57    & 45.48    & 77.26    & 61.37   & 53.78   \\
F-ConvNet\cite{fconvnet}           & L \& I             & 67.89   & 91.51  & 85.84  & 76.11  & 57.04     & 48.96    & 44.33    & \textbf{84.16}    & 68.88   & 60.05   \\ \hline
MMF\cite{liang2019multi}          & L, I  \& M      	& N/A  	& 93.67  & 88.21  & 81.99  & N/A       & N/A      & N/A      & N/A      & N/A     & N/A     \\ \hline
LaserNet\cite{lasernet}     		& L		         & N/A  	& 79.19. & 74.52  & 68.45   & N/A       & N/A      & N/A      & N/A      & N/A     & N/A     \\
SECOND\cite{second}      		& L                    &  61.61   & 89.39  & 83.77  & 78.59  & 55.99     & 45.02    & 40.93    & 76.5     & 56.05   & 49.45   \\
PointPillars\cite{pointpillars}  	& L                    & 65.98    & 90.07  & 86.56  & 82.81  & 57.60     & 48.64    & 45.78    & 79.90    & 62.73   & 55.58   \\
STD\cite{yang2019std}          	& L                    & 68.38    & \textbf{94.74}  & \textbf{89.19}  & \textbf{86.42}  & 60.02     & 48.72    & 44.55    & 81.36    & 67.23   & 59.35   \\
PointRCNN\cite{pointrcnn}    	& L                    & 66.92    & 92.13  & 87.39  & 82.72  & 54.77     & 46.13    & 42.84    & 82.56    & 67.24   & 60.28   \\ \hline
Painted PointRCNN  		& L \& I             & \textbf{69.86}	& 92.45  & 88.11  & 83.36  & 58.70     & 49.93    & 46.29    & 83.91    & \textbf{71.54}   & \textbf{62.97}    \\ \hdashline
Delta	 					& $\Delta$ I           & \greenbf{+2.94} &	\greenbf{+0.32} & \greenbf{+0.72} & \greenbf{+0.64} & \greenbf{+3.93} & \greenbf{+3.80}  & \greenbf{+3.45} & \greenbf{+1.35} & \greenbf{+4.30} &	\greenbf{+2.69}     \\ \hline
\end{tabular}
\vspace{1mm}
\caption{
Results on the KITTI test BEV detection benchmark.
We see that Painted PointRCNN sets a new state of the art (\textbf{69.86 mAP}) in BEV detection performance.
The modalities are lidar (L), images (I), and maps (M).
The delta is the difference due to Painting, ie Painted PointRCNN minus PointRCNN.
The corresponding 3D results are included in \tableref{table:res_3d} in the Supplementary Material.
}
\label{table:res_bev}
\end{table*}

Here we provide details on the painting algorithm.
Each point in the lidar point cloud is ($x$, $y$, $z$, $r$) or ($x$, $y$, $z$, $r$, $t$) for KITTI and nuScenes respectively, where $x$, $y$, $z$ are the spatial location of each lidar point, $r$ is the reflectance, and $t$ is the relative timestamp of the lidar point (applicable when using multiple lidar sweeps~\cite{nuscenes}).
The lidar points are transformed by a homogenous transformation followed by a projection into the image.
For KITTI this transformation is given by $T_{\rm{camera}\leftarrow\rm{lidar}}$.
The nuScenes transformation requires extra care since the lidar and cameras operate at different frequencies.
The complete transformation is:
\begin{equation*}
T=
T_{(\rm{camera}		\leftarrow		\rm{ego})}
T_{\rm{(ego_{t_c}}	\leftarrow		\rm{ego_{t_l}})}
T_{(\rm{ego}	\leftarrow		\rm{lidar})}
\end{equation*}
with transforms: \lidar frame to the ego-vehicle frame; ego frame at time of lidar capture, $t_l$, to ego frame at the image capture time, $t_c$; and ego frame to camera frame.
Finally, the camera matrix, $M$, projects the points into the image.

The output of the segmentation network is $C$ class scores, where for KITTI $C=4$ (car, pedestrian, cyclist, background) and for nuScenes $C=11$ (10 detection classes plus background).
Once the lidar points are projected into the image, the segmentation scores for the relevant pixel, ($h$, $w$), are appended to the lidar point to create the painted lidar point.
Note, if the field of view of two cameras overlap, there will be some points that will project on two images simultaneously and we randomly choose the segmentation score vector from one of the two images.
Another strategy can be to choose the more discriminative score vector by comparing their entropies or the margin between the top two scores. However, we leave that for future studies.

\subsection{Lidar Detection}
\label{sec:pointpillars}

The decorated point clouds can be consumed by any lidar network that learns an encoder, since PointPainting just changes the input dimension of the lidar points.
PointPainting can also be utilized by lidar networks with hand-engineered encoder~\cite{pixor, complexyolo}, but requires specialized feature engineering for each method.
In this paper, we demonstrate that PointPainting works with three different lidar detectors: PointPillars~\cite{pointpillars}, VoxelNet~\cite{voxelnet,second}, and PointRCNN~\cite{pointrcnn}.
These are all state of the art lidar detectors with distinct network architectures: single stage (PointPillars, VoxelNet) vs two stage (PointRCNN), and pillars (PointPillars) vs voxels (VoxelNet) vs point-wise features (PointRCNN).
Despite these different design choices, all lidar networks benefit from PointPainting (\tableref{table:point_delta}).
Note that we were as inclusive as possible in this selection, and to the best of our knowledge, these represent all of the top KITTI detection leaderboard methods that have public code.


\section{Experimental setup}
\squeeze
In this section we present details of each dataset and the experimental settings of PointPainting.

\subsection{Datasets}
\label{section:dataset}
\squeeze
We evaluate our method on the KITTI and nuScenes datasets.

\mypar{KITTI.}
The KITTI dataset~\cite{kitti} provides synced \lidar point clouds and front-view camera images.
It is relatively small with 7481 samples for training and 7518 samples for testing.
For our test submission, we created a minival set of 784 samples from the training set and trained on the remaining 6733 samples.
The KITTI object detection benchmark requires detection of cars, pedestrians, and cyclists.
Ground truth objects were only annotated if they are visible in the image, so we follow the standard practice~\cite{mv3d, voxelnet} of only using \lidar points that project into the image.

\mypar{nuScenes.}
The nuScenes dataset~\cite{nuscenes} is larger than the KITTI dataset (7x annotations, 100x images).
It it annotated with 3D bounding boxes for 1000 20-second scenes at 2Hz resulting in 28130 samples for training, 6019 samples for validation and 6008 samples for testing.
nuScenes comprises the full autonomous vehicle data suite: synced \lidar, cameras and radars with complete 360 coverage;
in this work, we use the \lidar point clouds and RGB images from all 6 cameras.
The 3D object detection challenge evaluates the performance on 10 classes: cars, trucks, buses, trailers, construction vehicles, pedestrians, motorcycles, bicycles, traffic cones and barriers.
Further, the dataset has an imbalance challenge with cars and pedestrians most frequent, and construction vehicles and bicycles least frequent.

\begin{table*}[]
\small
\center
\begin{tabular}{|c|c|c|c|c|c|c|c|c|c|c|c|}
\hline
Methods       						& mAP 	& Car      	& Truck         	& Bus           	& Trailer        	& Ctr. Vhl.  	& Ped.     		& Motorcycle    	& Bicycle 		& Tr. Cone       	& Barrier       	\\ \hline \hline
PointPillars~\cite{pointpillars,nuscenes}     & 30.5	& 68.4      & 23.0         	& 28.2          	& 23.4          	& 4.1      		& 59.7          	& 27.4          	& 1.1             	& 30.8          	& 38.9          	\\ \hline
PointPillars+   						& 40.1	& 76.0     	& 31.0    		& 32.1		& 36.6 		& 11.3      		& 64.0 		& 34.2		& 14.0		& 45.6         	& 56.4          	\\ \hline
Painted PointPillars+ 				& 46.4	& 77.9      & 35.8           	& 36.1 		& 37.3 		& 15.8          	& 73.3 		& 41.5		& 24.1		& 62.4         	& 60.2  		\\ \hdashline
Delta				 				& \greenbf{+6.3} 	& \greenbf{+1.9} 	& \greenbf{+4.8}           	& \greenbf{+3.9} 		& \greenbf{+0.7} 		& \greenbf{+4.5}          	& \greenbf{+9.3} 		& \greenbf{+7.3}		& \greenbf{+10.1}		& \greenbf{+16.8}         	& \greenbf{+3.8}		\\ \hline
\end{tabular}
\vspace{1mm}
\caption{
Per class nuScenes performance.
Evaluation of detections as measured by average precision (AP) or mean AP (mAP) on nuScenes test set.
Abbreviations: construction vehicle (Ctr. Vhl.), pedestrian (Ped.), and traffic cone (Tr. Cone).
}
\label{table:nuscenes_per_class}
\end{table*}

\squeeze
\subsection{Semantics Network Details}
\label{sec:seg_net}
\squeeze

Here we provide more details on the semantics networks.

\mypar{KITTI.}
For experiments on KITTI~\cite{kitti}, we used the DeepLabv3+ network\footnote{https://github.com/NVIDIA/semantic-segmentation}.
The network was first pretrained on Mapillary~\cite{mapillary}, then finetuned on Cityscapes~\cite{cityscapes}, and finally finetuned again on KITTI pixelwise sem. seg.~\cite{kitti}.
Note that the class definition of cyclist differs between KITTI sem. seg. and object detection: in detection a cyclist is defined as rider $+$ bike, while in sem. seg. a cyclist is defined as only the rider with bike a separate class.
There was therefore a need to map bikes which had a rider to the cyclist class, while supressing parked bikes to background.
We did this after painting by mapping all points painted with the bike class within a $1m$ radius of a rider to the cyclist class; the rest to background.

\mypar{nuScenes.}
There was no public semantic segmentation method available on nuScenes so we trained a custom network using the nuImages dataset.\footnote{We used an early access version; https://www.nuscenes.org/images.}
nuImages consists of $100$k images annotated with 2D bounding boxes and segmentation labels for all nuScenes classes.
The segmentation network uses a ResNet~\cite{resnet} backbone to generate features at strides $8$ to $64$ for a FCN~\cite{fcn} segmentation head that predicts the nuScenes segmentation scores.

\squeeze
\subsection{Lidar Network Details}
\squeeze

We perform experiments using three different lidar networks: PointPillars~\cite{pointpillars}, VoxelNet~\cite{voxelnet,second}, and PointRCNN~\cite{pointrcnn}.
The fusion versions of each network that use PointPainting will be referred to as being painted (e.g. Painted PointPillars).

\mypar{KITTI.}
We used the publicly released code for PointPillars\footnote{https://github.com/nutonomy/second.pytorch}, VoxelNet\footnote{https://github.com/traveller59/second.pytorch} and PointRCNN\footnote{https://github.com/sshaoshuai/PointRCNN} and decorate the point cloud with the sem. seg. scores for $4$ classes.
This changes the original decorated point cloud dimensions from $9\rightarrow13$, $7\rightarrow11$, and $4\rightarrow8$ for PointPillars, VoxelNet, and PointRCNN respectively.
For PointPillars, the new encoder has $(13, 64)$ channels, while for VoxelNet it has $(11, 32), (64, 128)$ channels.
The $8$ dimensional painted point cloud for PointRCNN is given as input to both the encoder and the region pooling layer.
No other changes were made to the public experimental configurations.

\mypar{nuScenes.}
We use PointPillars for all nuScenes experiments.
This requires changing the decorated point cloud from $7\rightarrow18$, and the encoder has $(18, 64)$ channels now.

In order to make sure the effect of painting is measured on a state of the art method, we made several improvemnts to the previously published PointPillars setup~\cite{nuscenes} boosting the mAP by 10\% on the nuScenes bechmark (\tableref{table:nuscenes_test}).
We refer to this improved baseline as PointPillars+. The changes are inspired by~\cite{megvii2019nuscenes} and comprise modifying pillar resolution, network architecture, attribute estimation, sample weighting, and data augmentation.
First, the pillar resolution was reduced from 0.25 m to 0.2 m to allow for better localization of small objects.
Second, the network architecture was changed to include more layers earlier in the network.
Third, neither PointPillars nor PointPillars+ predict attributes, instead the attribute estimation heuristic was improved.
Rather than using the most common attribute for each class, the predicted velocities and heights of each box are used to better estimate each attribute.
Fourth, to reduce the class imbalance during training, a sample based weighting method was used where each sample was weighted according to the number of annotations in the sample.
Fifth, the global yaw augmentation was changed from $\pi$ to $\pi/6$.


\section{Results} \label{sec:results}
\squeeze

In this section, we present PointPainting results on the KITTI and nuScenes datasets and compare to the literature.

\subsection{Quantitative Analysis}

\subsubsection{KITTI}
All detection results are measured using the official KITTI evaluation detection for bird's-eye view (BEV) and 3D.
The BEV results are presented here while the 3D results are included in the Supplementary Material.
The KITTI dataset is stratified into easy, moderate, and hard difficulties, and the official KITTI leaderboard is ranked by performance on moderate average precision (AP).

\mypar{Validation Set}
First, we investigate the effect of PointPainting on three leading lidar detectors.
Fig.~\ref{fig:point_painting_delta} and \tableref{table:point_delta} demonstrate that PointPainting improves the detection performance for PointPillars~\cite{pointpillars}, VoxelNet~\cite{voxelnet,second}, and PointRCNN~\cite{pointrcnn}.
The PointPainting semantic information led to a widespread improvement in detection: 24 of 27 comparisons ($3~\rm{experiments} \times 3 ~\rm{classes} \times 3 ~\rm{strata}$) were improved by PointPainting.
While the greatest changes were for the more challenging scenarios of pedestrian and cyclist detection, most networks even saw an improvement on cars.
This demonstrates that the utility of PointPainting is independent of the underlying lidar network.

\mypar{Test Set}
Here we compare PointPainting with state of the art KITTI test results.
The KITTI leaderboard only allows one submission per paper, so we could not submit all Painted methods from \tableref{table:point_delta}.
While Painted PointPillars performed better than Painted PointRCNN on the val set, of the two only PointPillars has public code for nuScenes.
Therefore, to establish the generality of PointPainting, we chose to submit Painted PointPillars results to nuScenes test, and use our KITTI submission on Painted PointRCNN.

As shown in \tableref{table:res_bev}, PointPainting leads to a robust improvement on the test set for PointRCNN: the average precision increases for every single class across all strata.
Painted PointRCNN establishes new state of the art performance on mAP and cyclist AP.

Based on the consistency of Painted PointRCNN improvements between val and test ($+2.73$ and $+2.94$ respectively), and the generality of PointPainting (\tableref{table:point_delta}), it is reasonable to believe that other methods in \tableref{table:res_bev} would decidedly improve with PointPainting.
The strength, generality, robustness, and flexibility of PointPainting suggests that it is the leading method for image-lidar fusion.

\begin{table}[]
\small
\center
\begin{tabular}{|c|c|c|c|}
\hline
Method									& Modality	 		& NDS	& mAP   	\\ \hline \hline
MonoDis~\cite{monodis}						& Images			& 38.4	& 30.4	\\ \hline
PointPillars~\cite{pointpillars,nuscenes}  			& Lidar			& 45.3 	& 30.5 	\\ \hline
PointPillars+    								& Lidar 			& 55.0 	& 40.1 	\\ \hline
Painted PointPillars+    						& Lidar \& Images 	& 58.1 	& 46.4	\\ \hline
MEGVII~\cite{megvii2019nuscenes}    			& Lidar 			& 63.3  	& 52.8 	\\ \hline
\end{tabular}
\vspace{1mm}
\caption{
nuScenes test results.
Detection performance is measured by nuScenes detection score (NDS)~\cite{nuscenes} and mean average precision (mAP).
}
\label{table:nuscenes_test}
\end{table}

\subsection{nuScenes}
To establish the versatility of PointPainting, we examine Painted PointPillars results on nuScenes.
As a first step, we strengthened the lidar network baseline to PointPillars+.
Even with this stronger baseline, PointPainting increases mean average precision (mAP) by $+6.3$ on the test set (\tableref{table:nuscenes_test}).
Painted PointPillars+ is only beat by MEGVII's lidar only method on nuScenes.
However, MEGVII's network~\cite{megvii2019nuscenes} is impractical for a realtime system since it is an extremely large two stage network that requires high resolution inputs and uses multi-scale inputs and ensembles for test evaluation.
Therefore, Painted PointPillars+ is the leading realtime method on nuScenes.

The detection performance generalized well across classes with every class receiving a boost in AP from PointPainting (\tableref{table:nuscenes_per_class}).
In general, the worst performing detection classes in PointPillars+ benefited the most from painting, but there were exceptions.
First, traffic cones received the largest increase in AP ($+16.8$) despite already having robust PointPillars+ detections.
This is likely because traffic cones often have very few lidar points on them, so the additional information provided by semantic segmentation is extremely valuable.
Second, trailer and construction vehicles had lower detection gains, despite starting from a smaller baseline.
This was a consequence of the segmentation network having its worst recall on these classes (overall recall of $72\%$, but only $39\%$ on trailers and $40\%$ on construction vehicles; see Supplementary Material for details).
Finally, despite a baseline of $76$ AP, cars still received a $+1.9$ AP boost, signaling the value of semantic information even for classes well detected by lidar only.

\begin{figure}
\begin{center}
\includegraphics[width = 8.3cm]{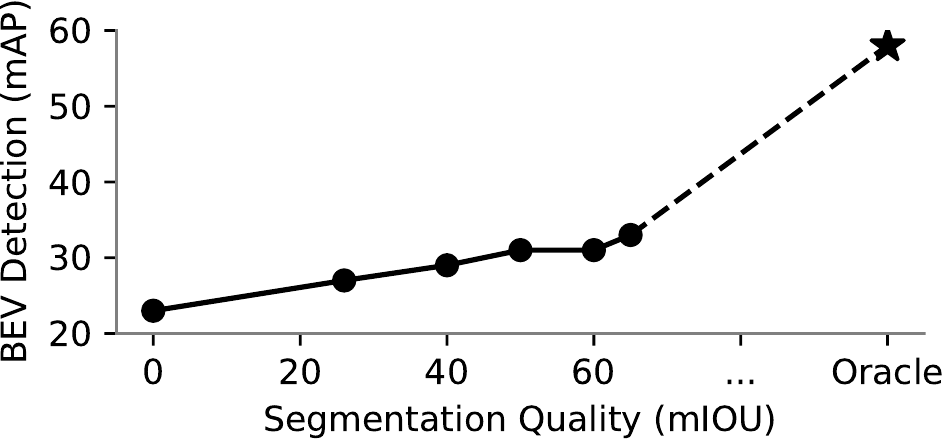}
\end{center}
\vspace{-4.9mm}
\caption{
PointPainting dependency on segmentation quality.
The Painted PointPillars detection performance, as measured by mean average precision (mAP) on the val split, is compared with respect to the quality of semantic segmentation network used in the painting step, as measured by mean intersection over union (mIoU).
The oracle uses the 3D bounding boxes as semantic segmentation.
}
\label{figure:segmentation_quality}
\end{figure}

\subsection{Qualitative Analysis}
Here we give context to the evaluation metrics with some qualitative comparisons in Fig.~\ref{fig:kitti_visualize} using Painted PointPillars, the best performing network on KITTI val set.
In Fig.~\ref{fig:kitti_visualize} A, original PointPillars correctly detects the cars, but misses a cyclist.
The painted point cloud resolves this and the cyclist is detected. It also yields better orientation estimates for the vehicles.
A common failure mode of lidar based methods is confusion between pedestrians and poles (Fig. ~\ref{fig:ped_vs_pole}).
As expected, PointPainting can help resolve this (Fig.~\ref{fig:kitti_visualize} B).
Fig.~\ref{fig:kitti_visualize} C suggests that the lidar detection step can correct incorrect painting.
The loose segmentation masks in the image correctly paint nearby pedestrians, but extra paint also gets splattered onto the wall behind them.
Despite this incorrect semantic information, the network does not predict false positive pedestrians.
This leaves unanswered the precise characteristics of sem. seg. (e.g. precision vs recall) to optimize for PointPainting.
In Fig.~\ref{fig:kitti_visualize} D, Painted PointPillars predicts two false positive cyclists on the left because of two compounding mistakes.
First, the sem. seg. network incorrectly predicts pedestrians as riders as they are so close to the parked bikes.
Next, the heuristic that we used to resolve the discrepancy in the cyclist definition between detection and segmentation annotations (See Section \ref{sec:seg_net}) exacerbated the problem by painting all bikes with the cyclist class.
However, throughout the rest of the crowded scene, the painted points lead to better oriented pedestrians, fewer false positives, and better detections of far away cars.


\begin{figure}
\begin{center}
\includegraphics[width = \columnwidth]{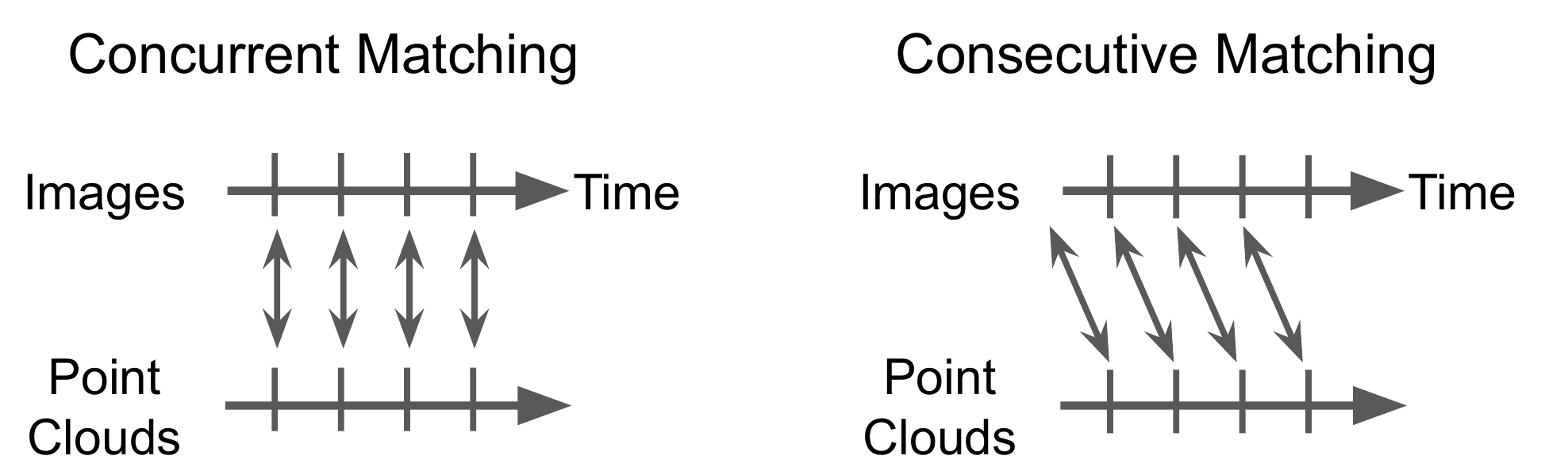}
\end{center}
\vspace{-3.3mm}
\caption{
Reducing latency by pipelining.
A Painted lidar network requires both point clouds and images.
Using the most recent image (Concurrent Matching) adds latency since the lidar network must wait for the image segmentation results.
This latency can be minimized using pipelining if the Painted network uses the segmentation mask of previous images (Consecutive Matching).
Using consecutive matching, we found that Painted PointPillars only adds a latency of 0.75 ms over the original PointPillars architecture.
See Supplementary Material for further details.
}
\squeeze
\label{figure:timing_schematic}
\end{figure}

\section{Ablation Studies}

Here we perform ablation studies on the nuScenes dataset.
All studies used the Painted PointPillars architecture and were trained for a quarter of the training time as compared to the test submissions.
Using the one-cycle optimizer~\cite{onecycle}, we achieved 33.9 mAP and 46.3 NDS on the nuScenes val set as opposed to 44.85 mAP and 57.34 NDS for full training of Painted PointPillars+.

\subsection{Dependency on Semantics}

\mypar{Quality.}
In PointPainting, the lidar points are fused with the semantic segmentation of the image.
We investigate the impact of the semantic segmentation quality on the final detection performance.
Using nuScenes, we generate a series of sem. seg. networks with varying segmentation quality by using multiple intermediate checkpoints from training.
As shown in Fig.~\ref{figure:segmentation_quality}, improved sem. seg. (as measured by mean IOU), leads to improved 3D object detection.

For an upper bound, we include an ``oracle" which uses the ground truth 3D boxes to paint the lidar points.
This significantly improves the detection performance ($+27$ mAP), which demonstrates that advances in semantic segmentation would radically boost 3D object detection.

Using the oracle doesn't guarantee a perfect mAP because of several limitations.
First, the ground truth bounding box can contain irrelevant points (e.g. from the ground).
Second, nuScenes annotates all objects that contain a single lidar point.
Turning one lidar point into an accurate, oriented 3D bounding box is difficult.
Third, we trained it for the same total time as the other ablation studies, but it would probably benefit from longer training.
Finally, PointPillars' stochastic sampling of the point cloud could significantly filter, or eliminate, the points that contain semantic information if the ground truth object contains only a few points.

\mypar{Scores vs Labels.}
We investigate the effect of the segmentation prediction format on detection performance.
To do so we convert the segmentation scores to a one hot encoding, effectively labelling each pixel as the class with the highest score.
When using the labels instead of scores, the NDS was unchanged and the mAP was, surprisingly, $+0.4$ higher.
However, the gains are marginal and within the noise of training.
We also hypothesize that for future studies, a combination of calibrated~\cite{calibration} segmentation scores and a larger PointPillars encoder would perform better.

Comparing these results with the segmentation quality ablation suggests that future research focus more on improving segmentation quality and less on representation.

\begin{table}[]
\small
\center
\begin{tabular}{|c|c|c|c|}
\hline
Method					& Matching		& NDS	& mAP	\\ \hline \hline
Painted PointPillars    		& Concurrent		& 46.3 	& 33.9 	\\ \hline
Painted PointPillars   		& Consecutive 		& 46.4  	& 33.9 	\\ \hline
\end{tabular}
\vspace{1mm}
\caption{
Time delay analysis.
Painted PointPillars results on nuScenes when using concurrent matching (which incurs latency), or consecutive matching (which allows real-time pipelining) as shown in \figref{figure:timing_schematic}.
The use of the previous image minimizes latency without any drop in detection performance.
}
\label{table:time_delay}
\end{table}

\subsection{Sensitivity to Timing}
\label{timing_sensitivity}

We investigate the sensitivity of the lidar network to delays in semantic information.
In the simplest scenario, which we used in all previous results, each point cloud is matched to the most recent image (Concurrent Matching - Fig.~\ref{figure:timing_schematic}).
However, this will introduce a latency in a real time system as the fusion step will have to wait for the image based sem. seg. scores.
To eliminate the latency, the sem. seg. scores of the previous image can be pipelined into the lidar network (Consecutive Matching - Fig.~\ref{figure:timing_schematic}).
This involves an ego-motion compensation step where the \lidar pointcloud is first transformed to the coordinate system of the ego-vehicle in the last frame followed by a projection into the image to get the segmentation scores.
Our experiments suggest that using the previous images does not degrade detection performance (\tableref{table:time_delay}).
Further, we measure that PointPainting only introduces an additional latency of 0.75 ms for the Painted PointPillars architecture (see Supplementary Material for details).
This demonstrates that PointPainting can achieve high detection performance in a realtime system with minimal added latency.


\squeeze
\section{Conclusion} \label{sec:conclusion}
\squeeze

In this paper, we present PointPainting, a novel sequential fusion method that paints lidar point clouds with image based semantics.
PointPainting produces state of the art results on the KITTI and nuScenes challenges with multiple different lidar networks.
The PointPainting framework is flexible and can combine the outputs of any segmentation network with any lidar network.
The strength of these results and the general applicability demonstrate that PointPainting is the leading architecture when fusing image and lidar information for 3D object detection.

\squeeze
\section{Acknowledgements}
\squeeze

We thank Donghyeon Won, Varun Bankiti and Venice Liong for help with the semantic segmentation model for nuScenes, and
Holger Caesar for access to nuImages.

{\small
\bibliographystyle{ieee}
\bibliography{../references}
}

\appendix

\clearpage

\begin{minipage}[t][2.9cm]{1\textwidth}
\centering
\Large \bf \title \par
PointPainting: Sequential Fusion for 3D Object Detection\\
\vspace{+4mm}
Supplementary Material\\

\end{minipage}

\section{PointPainting: 3D results}

In this section, we present 3D results of PointPainting on the KITTI validation and test sets.

\mypar{Validation Set}
Similar to \bev results (\tableref{table:point_delta}), we see that PointPainting substantially improves 3D detection performance on the validation set.
As seen in \tableref{table:point_delta_3d}, 23 out of 27 comparisons (3 experiments x 3 classes x 3 strata) were improved by PointPainting.

\mypar{Test Set}
In the test set (\tableref{table:res_3d}), we observe that PointPainting consistently improves 3D detection results of PointRCNN on the pedestrians and cyclists classes across all difficulty strata (easy, medium and hard).
However, we see that the 3D results on the car class drops substantially.
We think this could be because of overfitting on our minival set which was very small (see Section \ref{section:dataset}).

\section{PointPainting Latency}

In Section \ref{timing_sensitivity}, we concluded that Consecutive matching (see \figref{figure:timing_schematic}) can minimize the latency introduced by PointPainting without any drop in detection performance.
Here we provide a detailed breakdown of the latency introduced by PointPainting in the case of Consecutive matching.

\mypar{Projection}
This step involves transforming the pointcloud to the coordinate system of the ego-vehicle in the previous frame followed by a projection into the camera images to get the segmentation scores.
This operation only adds a latency of 0.15 ms.

\mypar{Encoding}
The Painted PointPillars encoder operates on an 18 dimensional decorated pointcloud as opposed to the 7 dimensional pointcloud in the original PointPillars architecture.
We measure the runtimes for both the encoders in TensorRT and find that PointPainting adds an additional latency of 0.6 ms in the encoding stage.\\
\\
Thus, Painted PointPillars only introduces an additional latency of 0.75 ms over PointPillars when Consecutive matching is used.
This makes Painted PointPillars a strong candidate for realtime camera-lidar fusion.

\section{nuImages Semantic Segmentation}

Here we present some stats on the semantic segmentation network that we trained on the nuImages dataset.
The mean intersection over union (mIoU) on the validation set was 0.65.
The class-wise precision and recall on the validation set is shown in \tableref{table:seg_results}.
Our model performs the best on the car class and worst on the construction vehicle and trailer classes.

\begin{table}[]
\vspace{+35mm}
\begin{tabular}{|c|c|c|}
\hline
\textbf{Class}       & \textbf{Recall (\%)} & \textbf{Precision (\%)} \\ \hline
Car                  & 94                   & 89                      \\ \hline
Bus                  & 71                   & 92                      \\ \hline
Construction Vehicle & 40                   & 58                      \\ \hline
Trailer              & 39                   & 79                      \\ \hline
Truck                & 69                   & 76                      \\ \hline
Motorcycle           & 89                   & 87                      \\ \hline
Bicycle              & 58                   & 84                      \\ \hline
Pedestrian           & 80                   & 86                      \\ \hline
Barrier              & 81                   & 80                      \\ \hline
Traffic Cone         & 78                   & 84                      \\ \hline
\end{tabular}
\vspace{1mm}
\caption{
Class-wise Precision and Recall of the semantic segmentation network trained on the nuImages dataset.
}
\label{table:seg_results}
\end{table}

\begin{table*}[]
\small
\center
\begin{tabular}{| c | c || c | c | c || c | c | c || c | c | c |}
\hline
\multirow{2}{*}{Method}  & mAP  & \multicolumn{3}{|c||}{Car} & \multicolumn{3}{|c||}{Pedestrian} & \multicolumn{3}{|c|}{Cyclist} \\ \cline{2-11}
& Mod. & Easy   & Mod.   & Hard   & Easy      & Mod.     & Hard     & Easy     & Mod.    & Hard    \\ \hline \hline
PointPillars~\cite{pointpillars}    		& 66.96    		& 87.22  & 76.95  & 73.52     		& 65.37  & 60.66  & 56.51             			& 82.29  & 63.26  & 59.82      \\
Painted PointPillars 				& 69.03    		& 86.26  & 76.77  & 70.25        		& 71.50  & 66.15  & 61.03		        			& 79.12  & 64.18  & 60.79			\\ \hdashline
Delta 		 				& \green{+2.07}    		& \red{-0.96}     & \red{-0.18}	   & \red{-3.27}          	& \greenbf{+6.13}  & \greenbf{+5.49}  & \greenbf{+4.52}		& \red{-3.17}  & \green{+0.92}  & \green{+0.97}			\\ \hline \hline
VoxelNet~\cite{voxelnet,second}       & 67.12   	 	& 86.85  & 76.64  & 74.41    		& 67.79  & 59.84  & 52.38           			& 84.92  & 64.89  & 58.59        		\\
Painted VoxelNet     				& 68.01    		& 87.15  & 76.66  & 74.75        		& 68.57  & 60.93  & 54.01					& 85.61  & 66.44  & 64.15			\\ \hdashline
Delta			 				& \green{+0.89}    		& \green{+0.3}  & \green{+0.02}  & \green{+0.34}     & \green{+0.78}  & \green{+1.09}  & \green{+1.63}					& \green{+0.69}  & \green{+1.55}  & \greenbf{+5.56}			\\ \hline \hline
PointRCNN~\cite{pointrcnn}      	& 67.01    		& 86.75  & 76.05  & 74.30     		& 63.29  & 58.32  & 51.59             			& 83.68  & 66.67  & 61.92        		\\
Painted PointRCNN  			& 70.34   		& 88.38  & 77.74  & 76.76          	& 69.38  & 61.67  & 54.58					& 85.21  & 71.62  & 66.98  			\\ \hdashline
Delta 		 				& \greenbf{+3.33} &  \greenbf{+1.63}  & \greenbf{+1.69}  & \greenbf{+2.46}          		& \green{+6.09}  & \green{+3.35}  & \green{+2.29}					& \greenbf{+1.53}  & \greenbf{+4.95} & \green{+5.06}		\\ \hline
\end{tabular}
\vspace{1mm}
\caption{
PointPainting applied to state of the art \lidar based object detectors.
All lidar methods show an improvement in 3D mean average precision (mAP) of car, pedestrian, and cyclist on KITTI \emph{validation} set, moderate split.
}
\label{table:point_delta_3d}
\end{table*}

\begin{table*}[]
\small
\center
\begin{tabular}{| c | c | c || c | c | c || c | c | c || c | c | c |}
\hline
\multirow{2}{*}{Method} & \multirow{2}{*}{Modality}  & mAP  & \multicolumn{3}{|c||}{Car} & \multicolumn{3}{|c||}{Pedestrian} & \multicolumn{3}{|c|}{Cyclist} \\ \cline{3-12}
                        				&                      & Mod. & Easy   & Mod.   & Hard   & Easy      & Mod.     & Hard     & Easy     & Mod.    & Hard    \\ \hline \hline
MV3D\cite{mv3d}     			& L \& I             & N/A  	& 74.97  & 63.63  & 54.00  & N/A       & N/A      & N/A      & N/A      & N/A     & N/A     \\
AVOD-FPN\cite{avod}       	& L \& I             & 54.86     & 83.07  & 71.76  & 65.73  & 50.46     & 42.27    & 39.04    & 63.76    & 50.55   & 44.93   \\
F-PointNet\cite{frustum} 		& L \& I            	& 56.02    & 82.19  & 69.79  & 60.59  & 50.53     & 42.15    & 38.08    & 72.27    & 56.12   & 49.01   \\
F-ConvNet\cite{fconvnet}           & L \& I             & \textbf{61.61}    & 87.36  & 76.39  & 66.69  & 52.16     & 43.38    & 38.80    & \textbf{81.98}    & \textbf{65.07}   & \textbf{56.54}   \\ \hline
MMF\cite{liang2019multi}          & L, I  \& M      	& N/A  	& \textbf{88.40}  & 77.43  & 70.22  & N/A       & N/A      & N/A      & N/A      & N/A     & N/A     \\ \hline
PointPillars\cite{pointpillars}  	& L                    & 58.29    & 82.58  & 74.31  & 68.99   & 51.45   & 41.92    & \textbf{38.89}    & 77.10    & 58.65   & 51.92    \\
STD\cite{yang2019std}          	& L                    & 61.25    & 87.95  & \textbf{79.71}  & 75.09   & \textbf{53.29}   & \textbf{42.47}    & 38.35    & 78.69    & 61.59   & 55.30   \\
PointRCNN\cite{pointrcnn}    	& L                    & 57.94    & 86.96  & 75.64  & \textbf{70.70}   & 47.98   & 39.37    & 36.01    & 74.96    & 58.82   & 52.53   \\ \hline
Painted PointRCNN  		& L \& I             & 58.82	& 82.11  & 71.70  & 67.08   & 50.32   & 40.97    & 37.87    & 77.63    & 63.78   & 55.89    \\ \hdashline
Delta	 					& $\Delta$ I           & \greenbf{+0.88} &	\red{-4.85} & \red{-3.94} & \red{-3.62} & \greenbf{+2.34} & \greenbf{+1.6}  & \greenbf{+1.86} & \greenbf{+2.67} & \greenbf{+4.96} &	\greenbf{+3.36}     \\ \hline
\end{tabular}
\vspace{1mm}
\caption{
Results on the KITTI test 3D detection benchmark.
The modalities are lidar (L), images (I), and maps (M).
The delta is the difference due to Painting, ie Painted PointRCNN minus PointRCNN.
We don't include a few entries from \tableref{table:res_bev} because LaserNet\cite{lasernet} did not publish 3D results and SECOND\cite{second}, IPOD\cite{ipod} no longer have their entries on the public leaderboard since KITTI changed to a 40 point interpolated AP metric instead of 11.
}
\label{table:res_3d}
\end{table*}

\end{document}